# Segmentation and ABCD rule extraction for skin tumors classification

[1]Mahammed MESSADI, [2]Hocine CHERIFI and [1]Abdelhafid BESSAID

[1]*Biomedical Engineering Laboratory, Department of Electrical and Electronics, Technology Faculty, AbouBekrBelkaid, Tlemcen University - 13000*
m_messadi@yahoo.fr, a.bessaid@gmail.com
[2]*Laboratoire Electronique, Informatique et Image UMR CNRS 6306*
*Université de Bourgogne*
hocine.cherifi@u-bourgogne.fr

### *Abstract*

*During the last years, computer vision-based diagnosis systems have been widely used in several hospitals and dermatology clinics, aiming at the early detection of malignant melanoma tumor, which is among the most frequent types of skin cancer. In this work, we present an automated diagnosis system based on the ABCD rule used in clinical diagnosis in order to discriminate benign from malignant skin lesions. First, to reduce the influence of small structures, a preprocessing step based on morphological and fast marching schemes is used. In the second step, an unsupervised approach for lesion segmentation is proposed. Iterative thresholding is applied to initialize level set automatically. As the detection of an automated border is an important step for the correctness of subsequent phases in the computerized melanoma recognition systems, we compare its accuracy with growcut and mean shift algorithms, and discuss how these results may influence in the following steps: the feature extraction and the final lesion classification. Relying on visual diagnosis four features: Asymmetry (A), Border (B), Color (C) and Diversity (D) are computed and used to construct a classification module based on artificial neural network for the recognition of malignant melanoma. This framework has been tested on a dermoscopic database [16] of 320 images. The classification results show an increasing true detection rate and a decreasing false positive rate*.

## 1. Introduction

Melanoma is one of the most dangerous diseases, and its frequency is rising in many countries. The rising rate of skin cancer is a growing concern worldwide [1-2]. Mass screening for melanoma and other cutaneous malignancies has been advocated for early detection and effective treatment [3]. Thus, the development of a non-invasive imaging and analysis method could be beneficial in the early detection of cutaneous melanoma. Dermatologists use the ABCD rule (Asymmetry, Border, Colors, and Diameter) to characterize skin lesions [4-5-6-7]. The choice of this rule is based on dermatology criteria: shape, color and symmetry. Thus, there has been an increasing interest in computer-aided systems for the clinical diagnosis of melanoma as a support for dermatologists in different analysis steps, such as lesion boundary detection, extraction of the ABCD parameters and classification into different types of lesions.

Many works have been presented in this field in order to improve the early detection of melanomas. Most of these studies have suggested attributes that do not admit an accurate evaluation to differentiate benign lesions from malignant tumors. We can note that Lee [10] considers only the contour irregularity. This algorithm starts with a segmentation method, followed by a smoothing operation done by a fixed grain Gaussian filter and a growing standard deviation. We must also note that in the work of K.M. Clawson [11], the authors use the radial distribution of pigments along the contours in order to evaluate the asymmetry parameter. The sensitivity of the ABCD rule is reported to be between 59% and 88% in the works [8, 9]. In [12] the authors investigate melanoma diagnosis based on texture analysis and classified by ANN build up upon a database of 102 dermoscopic images (51 images for malignant melanoma and 51 images for benign nevi). Results show that their algorithm is able to classify malignant melanoma with 92 % accuracy of the test set. In [13] the statistical textural features extraction derived from GLCM for classification of skin tumors are used. The results of this study are consistent with theory that using photographic image instead of dermoscopic images is promising as it provides high accuracy rates. In [14] many segmentation techniques have been compared in order to identify the more accurate, and the authors discuss how these results may influence the feature extraction and the final lesion classification. In [15] the authors proposed a framework based on a combination of segmentation methods in order to develop an interface that can assist dermatologists in the diagnostic phase. The experiment uses 40 images containing suspicious melanoma skin cancer; the accuracy of the system reported is 92%. In this purpose, a variety of image segmentation methods have been proposed, such as GrowCut (GC) [17] and Mean shift (Ms) algorithms [18]. All these works demonstrate that the separation of lesion from background is a critical early step in the analysis of dermatoscopic imagery.

In this work we propose a new segmentation algorithm and characterization approach of specific attributes that can be used for computer-aided diagnosis of melanoma. The automatic diagnosis system is made of four main modules. The first module aims at removing artifacts on the original images such as hair that may influence the detection of skin lesion. The second module is a segmentation algorithm that identifies the lesion boundaries assuming that skin lesions are darker than healthy skin. The third module inspired by the ABCD rule based on visual diagnosis compute a set of four parameters Asymmetry (A), Border (B), Color (C) and Diversity (D) that allows to discriminate melanoma and benign lesions. Finally after extracting these features that characterize each image, a classification process is used to obtain a diagnosis of the imaged lesion.

To summarize, the main contribution of this work is twofold. First of all, we propose an unsupervised segmentation algorithm, based on thresholding [19] and on level-sets [20]. We show that this method is more efficient than prominent segmentation methods such as the GrowCut algorithm [17] and the Mean shift approach [18]. Upon comparison, the proposed method demonstrates good performance in achieving an automatic segmentation on various real skin data downloaded from public database [16].

Secondly we define and investigate a set of robust parameters based on the ABCD rule in order to discriminate efficiently melanoma and benign lesions. Regarding asymmetry, which is one of the clinical feature suggestive of malignancy, a set of techniques used for the detection of asymmetry are evaluated and a new method for quantifying asymmetry is proposed. Furthermore, we present a new method to measure the border irregularity. It relies on a hybrid approach, which consists on generating a circle whose shape is indicative of pixel distribution inside the lesion to quantify the border. Another objective of this work is the detection of the color information contained in the lesion, which is considered as an important feature used to identify skin tumors. For this part, we use texture parameters for the extraction of color information's. The last feature that we consider is the diameter because melanomas usually start with a diameter of more than 6–7 mm.

The rest of this paper is organized as follows: in section 2, we present the pre-processing step that we implemented in order to remove surrounding hair. In section 3, we recall briefly the GrowCut algorithm [17] and the Mean shift approach [18] together with the proposed segmentation approach. Furthermore we present the results of an extensive comparative evaluation of the proposed scheme. In section 4, we present the sequences of transformations that we propose to apply in order to measure the set of attributes (A: asymmetry, B: border, C: color and D: diameter) characterizing the image. In section 5 we present the experimental results of an artificial neural network classification process used to differentiate malignant tumors from benign lesions. We conclude the paper in section 6.

## 2. Pre-processing

Dermatologists can achieve an early detection of the skin tumor by studying the medical history of the patient, and also by examining the lesion in term of edge, shape, texture and color. Before such an examination, it is necessary to start with pre-processing and segmenting the skin tumor image. There are technical difficulties concerning the image segmentation in term of variations of the brightness, presence of artifacts (the hair) and the variability of edges. Therefore, filtering the noised image is necessary.

In this section we present a hair-removal algorithm that does not interfere with the tumor's texture [21], This approach restores the information occluded by hairs of any thickness. The algorithm is divided into three steps: (a) hair detection with the use of a derivative of Gaussian [22], (b) refinement by morphological techniques and (c) then hair repair by fast marching image inpainting [23] technique. The schematic of the proposed hair removal algorithm is displayed in fig. 1.

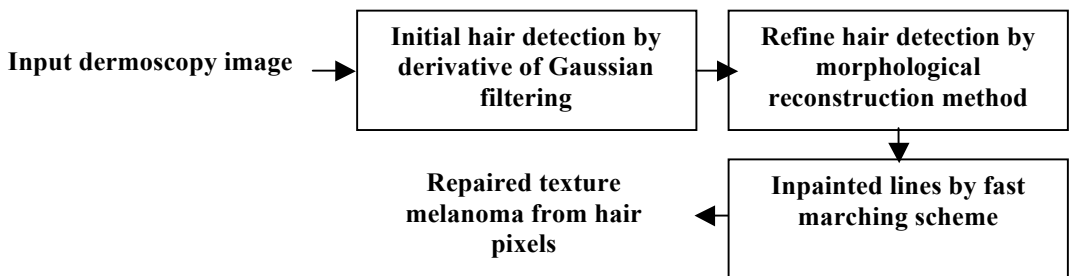

**Figure 1.** Systematic flow diagram of the proposed hair-occluded information repair algorithm.

Figure 2 is a typical example of the hair removing process. We can notice, according to fig. 2(c), that the images including thick hairs with color hue similar to the one of the lesion are removable by this method.

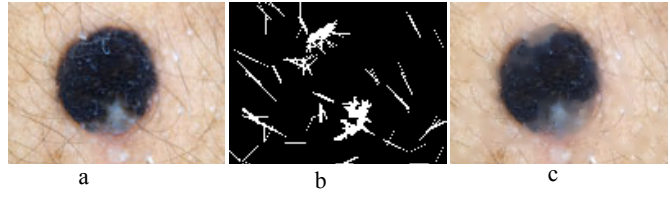
a        b        c
**Figure 2.** Hair removal by our method, (a) original image, (b) mask hairs,
(c) the image after hair removal

## 3. Tumors border detection

The quality of interpretation of a color image depends heavily on the segmentation process, which plays a major role in image processing and computer vision. It must achieve the difficult task of extracting useful information to locate and delineate the regions digital images. For this purpose, three different approaches (GrowCut, Mean shift and our approach) have been used for the segmentation of 320 dermatoscopic images selected randomly from the clinical database [16].

### 3.1 GrowCut method

This method utilizes statistical seed distributions to overcome the local bias seen in the traditional cellular automata framework [24]. We take an intuitive user interaction scheme - user specifies certain image pixels (seed pixels) that belong to objects that should be segmented from each other. The task is to assign labels to all other image pixels automatically, preferably achieving the segmentation result the user is expecting to get. This method uses cellular automaton to solve the pixel-labeling task. Commonly used neighborhood systems N is the von Neumann (equation 1) [17]:

$$N(p) = \{q \in Z^n : \|p - q\| \coloneqq \sum_{i=1}^{n} |p_i - q_i| = 1\} \qquad (1)$$

The cell state $S_p$ in our case is actually a triplet $(I_p, \theta_p, \vec{C_p})$ the Label $I_p$ of the current cell, 'strength' of the current cell $\theta_p$ and cell feature vector $\vec{C_p}$, defined by the image. Without loss of generality we will assume $\theta_p \in [0, 1]$. When user starts the segmentation by specifying the segmentation seeds, the seeded cells labels are set accordingly, while their strength is set to the seed strength value. This sets the initial state of the cellular automaton [17]. An example of an image segmentation obtained with the growcut algorithm is shown in figure 3.

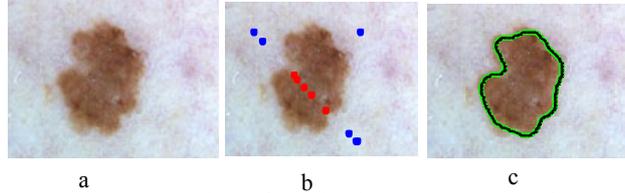
a        b        c
**Figure 3.** Segmentation of a color image. (a) source image, (b)
user-specified seeds, (c) segmentation results

### 3.2 Mean shift

Mean shift method is a non-parametric technique to examine a complex multi-modal feature space and to classify feature clusters. Size and shape of the region of interest are the only free parameters in this method. A two step sequence of discontinuity preserving filtering and mean shift clustering is utilized in this segmentation technique [25]. Starting from an initial estimate, the mean shift is calculated as below.

$$m_{h,G}(x) = \frac{\sum_{i=1}^{n} x_i g\left(\left\|\frac{x-x_i}{h}\right\|\right)^2}{\sum_{i=1}^{n} g\left(\left\|\frac{x-x_i}{h}\right\|\right)^2} - x \qquad (2)$$

Here, $x_i$ is the initial estimate of this iterative method, $g\left(\left\|\frac{x-x_i}{h}\right\|\right)$ can be considered as the kernel function which determines the weight of nearby points for re-estimation of the mean. This algorithm places $x \leftarrow m_{h,G}(x)$ and iterates until $m_{h,G}(x)$ converges to $x$. An example of an image segmentation using the mean shift algorithm is shown in figure 4.

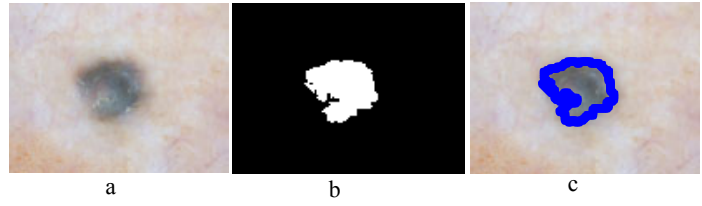

a             b             c

**Figure 4.** Segmentation of a color image. (a) source image,
(b) mask result, (c) segmentation results

### 3.3 The proposed method using unsupervised approach

For automatic border detection, our approach consists in two steps. In the first step, all images are rescaled to a standard size. In the second step, a segmentation algorithm based on thresholding and on level-sets is applied on the gray scale converted color image. The flow diagram of this unsupervised border detection method is shown in (Fig. 5).

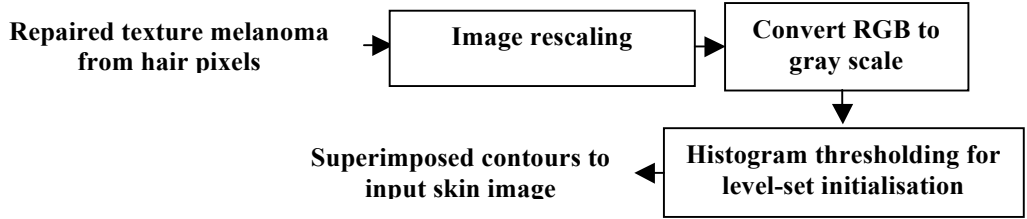

**Figure 5.** Flow diagram of the proposed unsupervised skin lesion border detection.

In particular, the evolution of Φ is totally determined by the numerical level set equation:

$$\begin{cases} \frac{\partial \Phi}{\partial t} + F \, |\nabla_\Phi| = 0 \\ \Phi(0, x, y) = \Phi_0(x, y) \end{cases} \quad (3)$$

Where $|\nabla_\Phi|$ denotes the normal direction, $\Phi_0(x, y)$ is the initial contour and $F$ represents the comprehensive forces [26]. The advancing force $F$ has to be regularized by an edge indication function $g$ in order to stop the level set evolution near the optimal solution:

$$g = \frac{1}{1+|\nabla(G_\sigma * I)|^2}, \quad (4)$$

Where $G_\sigma * I$ is the convolution of the image **I** with a smoothing Gaussian kernel $G_\sigma$. A popular formulation for level set segmentation is [27]:

$$\frac{\partial \Phi}{\partial t} = g|\nabla_\Phi|\left(\text{div}\left(\frac{\nabla_\Phi}{|\nabla_\Phi|}\right) + v\right), \quad (5)$$

Where div $\left(\nabla_\Phi / |\nabla_\Phi|\right)$ approximates mean curvature $k$ and $v$ is a customable balloon force.

$$\Phi_0(x, y) = \begin{cases} -C, & \Phi_0(x, y) < 0 \\ C, & \text{orherwise} \end{cases} \quad (6)$$

$$\Phi^{k+1}(x, y) = \Phi^k(x, y) + \tau[\mu\zeta(\Phi^k) + \xi(g, \Phi^k)]. \quad (7)$$

In general, the level set segmentation algorithm start by using an arbitrary boundary initialization corresponding to a binary region. The thresholding algorithm proposed in this paper, automates the initialization and the parameter configuration of the level set segmentation phase [28]. It is therefore convenient to initiate the level set function as:

$$\Phi_0(x, y) = -\varepsilon(0.5 - \mathbf{B}_k) \quad (8)$$

Where $\varepsilon$ is a constant regulating the Dirac function and $\mathbf{B}_k$ is a binary image obtained [28]. The Dirac function is then defined as follows:

$$\delta_\varepsilon(x) = \begin{cases} 0, & |x| > \varepsilon \\ \frac{1}{2\varepsilon}\left[1 + \cos\left(\frac{\pi x}{\varepsilon}\right)\right], & |x| \leq \varepsilon \end{cases} \quad (9)$$

Given the initial level set function $\Phi_0$ from thresholding as in Eq. (8), it is convenient to estimate the length $\ell$ and the area $\alpha$ by:

$$\ell = \int_I \delta(\Phi_0)\, dx\, dy; \qquad (10)$$

$$\alpha = \int_I H(\Phi_0)\, dx\, dy; \qquad (11)$$

Where the Heaviside function $H(\Phi_0)$ is:

$$H(\Phi_0) = \begin{cases} 1, & \Phi_0 \geq 0 \\ 0, & \Phi_0 < 0 \end{cases} \qquad (12)$$

The advantages of using this approach for skin lesion segmentation is that it has capabilities to separate heterogeneous objects, insensitive to noise, and automatic convergence along with control of overlapping contours into some extent.

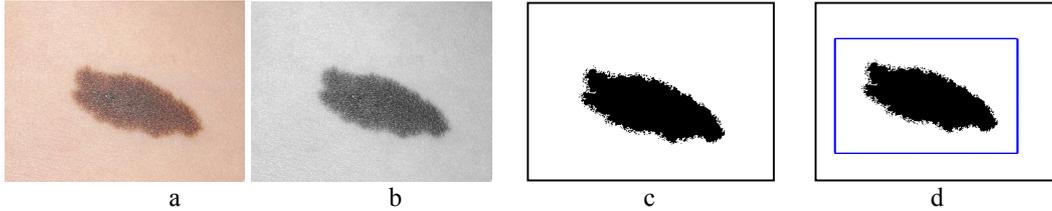

| a | b | c | d |

**Figure 6.** Level set initialization through adaptive thresholding : (a) original input RGB image after preprocessing steps, (b) convert into grey scale image, (c) initial tumors boundaries determine by thresholding, and (d) initialize contour.

A lesion is segmented with the initial detection of a lesion shape within a skin image by performing a thresholding method on the luminance values of the original color image. It is performed by calculating the mini–max intensity histograms on this luminance image. Finally, by using these mini–max intensity values, it computes threshold values used for detection of a lesion shape. Figure 6 illustrates this process. First we convert the RGB color image 6(a) into the grey scale image 6(b). By calculating the histogram thresholding, we obtain the segmented binary image as shown in Fig. 6(c). In this image, the tumors represent 0 intensity values, while a background is represented by 1 value. Therefore, we can initialize a level set curve closer towards the tumors as displayed in Fig. 6(d).

### 3.4 Comparison of Segmentation Methods Based on Experimental Results

We use for our experiments, 320 color images representing melanoma and benign lesions [16]. This clinical online database of dermoscopic and clinical view lesion images were obtained from various sources but most images came from the Department of Dermatology, Health Waikato New Zealand [16]. The images have been stored in the RGB color format of variable dimensions. The database is subdivided into six categories: 50 benign melanocytic lesions, 70 malignant melanoma lesions, 40 basal cell carcinoma lesions, 40 merkel cell carcinoma lesions, 90 seborrhoeic keratosis lesions and 30 non-melanocytic lesions. We implemented the pre-processing method to the image database in order to reduce the influence of small structures, hairs and bubbles,

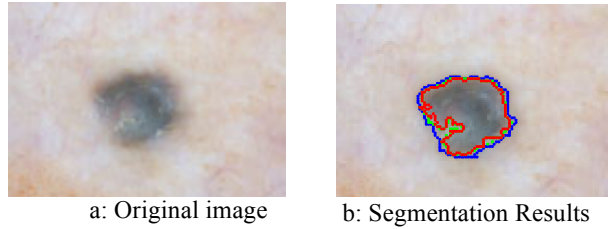

a: Original image    b: Segmentation Results

**Figure 7**. Segmentation of a dermatoscopic images by the Mean shift (red), GrowCut algorithm (green) and unsupervised approach (blue)

Segmentation has been performed using the GrowCut method (GC), Mean shift algorithm (Ms) and the proposed approach (UA). In order to compare the performances of the three segmentation methods presented above, we use the same image dataset [16] that has been manually segmented. The segmentation results are evaluated quantitatively using the classical error metric given by the following equation [29] [30] that measures the discrepancy between automatic and manual borders:

$$\text{Border Error} = \frac{\text{Area}(B \otimes A)}{\text{Area}(A)} \times 100\% \qquad (13)$$

Where A is a binary image such that all pixels inside the curves are produced by a clinical expert and B is the set of all lesion pixels labeled by a segmentation algorithms. $B \otimes A$ represents the differential segmentation obtained by the manual and automatic borders using XOR operation, and Area. ( . ) denotes the number of pixels in the binary images of A and $B \otimes A$.

Table 1 reports the performance of the three segmentation methods (Mean shift, GrowCut method and unsupervised approach) according to the clinical evaluation (manual method). The mean border error together with the results and a subjective evaluation by an expert are given. Indeed, each segmentation result has been evaluated by a dermatologist and rated in one of four possible labels: Vg-very good, G-good, Av-average, B-bad. Along with border error, table 1 presents the number of cases rated in each of the four classes by the different segmentation methods and the percentage of images rated very good. Results demonstrate that UA method outperforms both Mean Shift and Growcut. Note that the difference with Growcut is less pronounced. Globally the proposed method is adapted to this problem due to its simplicity, computational efficiency, and excellent performance on a variety of image domains. The result of the mean border error corroborates the expert ones. Indeed ,for this performance criterion, the best result is obtained by the UA with a score of 4.48% (Table 1).

**Table 1.** The performance of the three segmentation methods

|  | **Ms** | **GC** | **UA** |
|---|---|---|---|
| Vg | 207 | 260 | 280 |
| G | 43 | 34 | 20 |
| Av | 30 | 16 | 11 |
| B | 40 | 10 | 9 |
| Vg(%) | 64,68% | 81,25% | 87,5% |
| Border Error | 10.03 % | 07.02 % | 04.48% |

## 4. Feature Extraction for Skin Lesion Discrimination

This step aims to design a set of robust parameters that accurately describe each lesion and this in order to ensure that melanoma and benign lesions can be distinguished. We mimic the ABCD rule used by physicians in order to distinguish between different tumors [31].

### 4.1 Asymmetry index

Asymmetry (A) is one of the more important parameters used in differentiating malignant tumors from benign lesions. An asymmetry index based on the principal axes of the lesion, was proposed by Stoecker [32]. In this method, an asymmetry index is calculated from the smallest difference between the image area of the lesion and the image of the lesion reflected from the principal axis (Fig. 8). According to dermatologists, four axes are sufficient to determine the rate of symmetry (vertical, horizontal and two diagonal axes). To calculate this parameter, we determine the symmetry rotation of tumor through a 180° angle from first axe and second axe. Let A(x, y) be the initial surface, and B(x, y) the surface we obtained after symmetry rotation. The ratio between the intersection of A(x, y) surfaces and B(x, y) surfaces and their merging quantifies the recovery rate of the two surfaces, and therefore the degree of symmetry (equation 14). The calculation of this index is illustrated in Fig. 9, where the region in blue refers to the intersection of the surfaces and the external line defines their merging. The more the index approaches 1, the more the lesion will be considered as symmetrical.

$$IS = \frac{A \cap B}{A \cup B} \qquad (14)$$

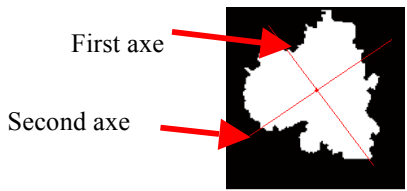
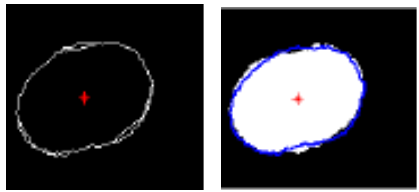

**Figure 8.** Principal axis illustration.   **Figure 9.** Example of a mask overlaid on its symmetrical (180° center rotation)

## 4.2 Border irregularity

We can also use the border irregularity in order to give an overview of the edge type that can be found. The irregularity parameter in a lesion is a very important factor when evaluating a malignant lesion. We use four special features to quantify irregular edges: compactness, fractal dimension, radial variance and extraction of small changes in the contour.

### 4.2.1 Index compact and fractal dimension

The compactness is implemented according to equation (15). It is evaluated on circles of different sizes. The compactness (c) of circle is equal to 1.

$$c = \frac{P^2}{4\pi a} \qquad (15)$$

Where p and a represent the perimeter and area of the lesion respectively. The lesion borders L are also evaluated using the fractal dimension D [33-34]. The lesion (L) is represented by a binary mask. Consequently the object is designated by 1, and the bottom by 0. The fractal dimension method is used to count the number of boxes N(r) contained in the border. The value D is estimated using two terms (r and N (r)) presented by equation (16) written as follows:

$$\log(N(r)) = D \times \log(r) + C^{ste} \qquad (16)$$

A circle of radius 100 pixels is used to test the fractal dimension (DF). The result obtained by the DF is equal to 99%, which signify that the error is 1% (the theoretical value of the DF is equal to 100%). Figures 10.a and 10.b represent the compactness and the DF for malignant and benign tumors, respectively.

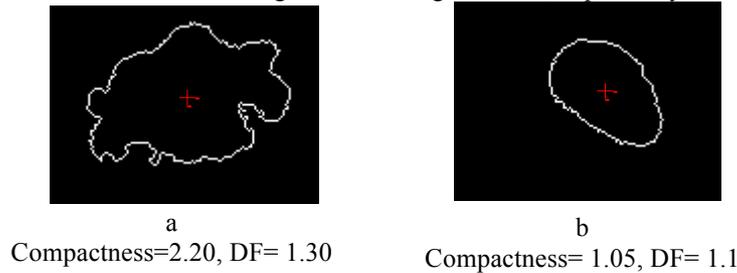

a
Compactness=2.20, DF= 1.30

b
Compactness= 1.05, DF= 1.1

**Figure 10.** Examples of applications, a: melanoma, b: benign tumor

### 4.2.2 Radial variance

A lesion with irregular border has a large variance in the radial distance. The border irregularity is estimated by the variance of the radial distance distribution [35]:

$$Ed = \frac{\frac{1}{P_L}\sum_{P \in C}(d(p,G)-m)^2}{m^2} \qquad (17)$$

Where m presents the average distance d between the boundary points and the centroid G. From the distance Ed, we draw a circle that represents the radial distribution of the tumor. To quantify the border parameter, we calculate the ratio between the area of the new circle and the surface of the tumor (Fig.11).

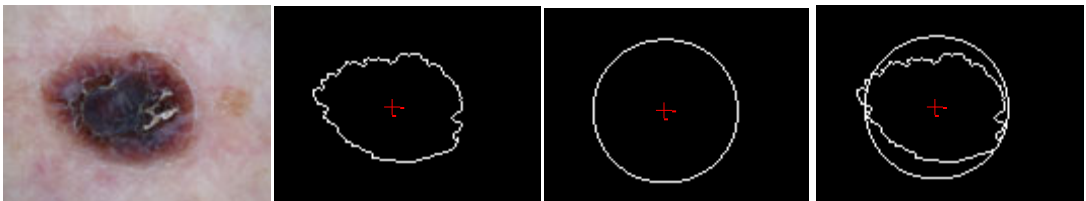

**Figure11.** Draw of the new circle, a: new circle, b: superposition of the two edges
(newcircle and the initial contour).

### 4.2.3 Our method for extraction of Small changes

Lesion irregularity is a very important factor for skin tumors evaluation. In this work, we use the Lee's algorithm [10] in order to quantify the irregularity of the contour. This algorithm starts by the boundary extraction in the image using equation 18.

$$L_0 = (x(t), y(t)) \quad (18)$$

In a second step, the initial contour $L_0$ is smoothed using a Gaussian filter (g). When σ is increased the border is more smoothed.

$$L(t, \sigma) = L_0(x(t), y(t)) \otimes g(t, \sigma) \quad \text{With} \quad g(t, \sigma) = \frac{1}{\sigma\sqrt{2\pi}} e^{-t^2/2\sigma^2}$$
$$= (X(t, \sigma), Y(t, \sigma)) \quad (19)$$

The image resulting is obtained by the convolution of the coordinates of the original image by a Gaussian filter (Fig. 12). As it is mentioned previously, when we increase the σ value, the nodulations are detected accurately.

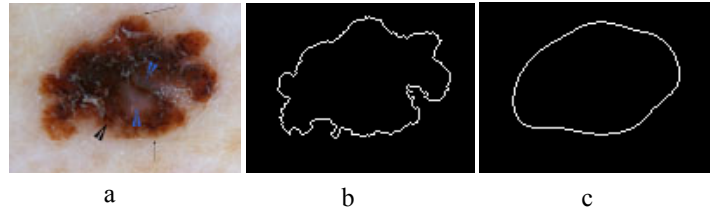

a        b        c

**Figure 12.** Image smoothing by a Gaussian filter, a: original image,
b: original contour, c: smooth contour after 12 iterations.

To compute the irregularity index, it is sufficient to calculate the ratio between initial and resulting images contours. In order to quantify the contour irregularity parameter, and to solve the problem of the smoothing operation, we propose in this paper a method that integrates the two algorithms used by Lee et al. [10] and Clawson et al [11]. This algorithm exploits the information contained in gray scale images. After gray scale inversion, for each radial path (from the lesion centroid to its boundary points) the mean gray scale value, $Av_i$, along that path is defined as:

$$Av_i = \frac{\sum_{k=1}^{N_i} r_{ik}}{N_i} \quad (20)$$

Where $r_{ik}$ is the value of pixel k along the radial path corresponding to boundary point i, and $N_i$ is equal to the number of pixels along the radial path from lesion centroid to boundary point i. A per-lesion indication of grayscale distribution can then be calculated as:

$$A_L = \frac{\sum_{i=1}^{N} Av_i}{N} \quad (21)$$

Where N is the number of lesion boundary points. Finally, for each boundary point, a Normalized color Distance (NCD) is derived using:

$$NCD = Av_i \left(\frac{100}{A_L}\right) \quad (22)$$

Our set of NCD values may then be used to generate a new contour, the shape of which is indicative of pigment distribution within the lesion.
So, ultimately our algorithm is summarized as follows:
For each image
- ❖ Smoothing the image
  - ➢ Calculation of the average along the path $A_L = \frac{\sum_{i=1}^{N} Av_i}{N}$ and $NCD = Av_i \left(\frac{100}{A_L}\right)$
  - ➢ Generate a new contour with a radius NCD
- ❖ Test operation
  - ➢ Calculate and draw a circle for each smoothed contour (using operations 1 and 2)
  - ➢ Calculate the tumor and circle compactness's (Ctumor, Ccircle))
  - ➢ Define a stopping criterion Si {( Ctumor, Ccircle< ε) } ( ε: the smallest possible difference)

$$I_R = \frac{Contour(C_{tumeur} \approx C_{cercle})}{Contour\ initial}$$

❖ End For

End For each image

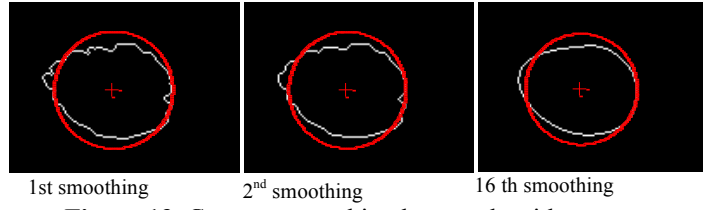

1st smoothing     2$^{nd}$ smoothing     16 th smoothing

**Figure 13.** Contour smoothing by our algorithm

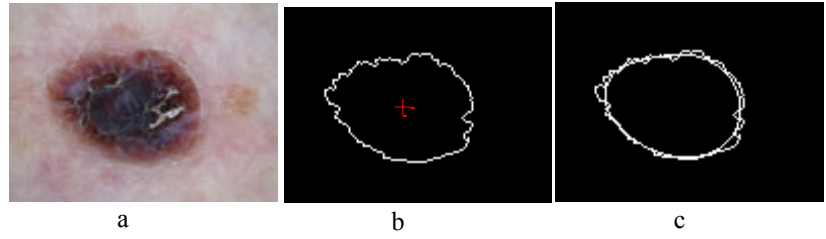

a     b     c

**Figure 14.** Contour smoothing, a: Initial image, b: Contour of the initial image,
c: Superposition of the two contours (original and smoothed).

Our methodology integrates the two previous algorithms in order to estimate the irregularity index. Figure 13 and figure 14 show the obtained results after the smoothing operation. At each iteration, we estimate the difference between compactness of the tumor and the compactness of the new obtained circle. In this example, after the 16$^{th}$ smoothing operation, the compactness's are nearly equal and the irregularity index (Ir) of this image was computed equal to 82.19%.

### 4.3 Color criterion

Our objective in this section is to detect the color information contained in the lesion. The methodology that we propose embraces texture and form parameters in order to achieve better classification results. Texture information is an important and efficient measure used to estimate the structure, orientation, roughness, or regularity of various regions in a set of images that enables us to distinguish between different objects [36]. In our work, we selected four parameters from the 14 proposed by Haralick [37]:

**Correlation:** This attribute evaluates the color distribution on the lesion.

$$Cor = \sum_{i=0}^{Ng} \sum_{j=0}^{Ng} \frac{(i-\mu_i)(j-\mu_j)}{\sigma^2} p(i,j) \qquad (23)$$

**Homogeneity:** This parameter indicates the grey levels uniformity measure in the image.

$$CH = \sum_{i=0}^{Ng} \sum_{j=0}^{Ng} \frac{p(i,j)}{1+|i-j|} \qquad (24)$$

**Energy:** This parameter presents a low value when the p(i, j) values are very close, and corresponds to a high value when some values are large and the other small.

$$E_n = \sum_{i=0}^{Ng} \sum_{j=0}^{Ng} [p(i,j)]^2 \qquad (25)$$

**Contrast:** This parameter measures the local variations present in the image.

$$Contr = \sum_{i=0}^{Ng} \sum_{j=0}^{Ng} [(i-j)^2] p(i,j) \qquad (26)$$

### 4.4 Diameter

The diameter (long axis of the lesion) is one of the ABCD criteria. The algorithm that we propose for calculating the diameter is presented as follows:
- ➢ Image segmentation (edge extraction)
- ➢ Determine the coordinates (x, y) of each pixel of the lesion perimeter
- ➢ Calculate the distance between each pair of points
- ➢ The maximum of these distances is the diameter

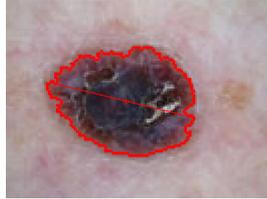

**Figure 15.** Original image with its diameter

## 5. Lesions classification and experimental results

We have seen that, in addition to the difficulty of standardizing the diagnostic criteria and the wide variability of the encountered structures, discrimination of certain types of lesion remains problematic. A classification system that allows tumors discrimination and analysis would be useful, especially for general practitioners who do not often observe melanomas. Such a system is introduced in figure 16, which presents a general methodology based on the extraction of pertinent parameters. In order to classify the tumor as melanoma or benign, a multilayer neural network with supervised learning algorithms is used [38].

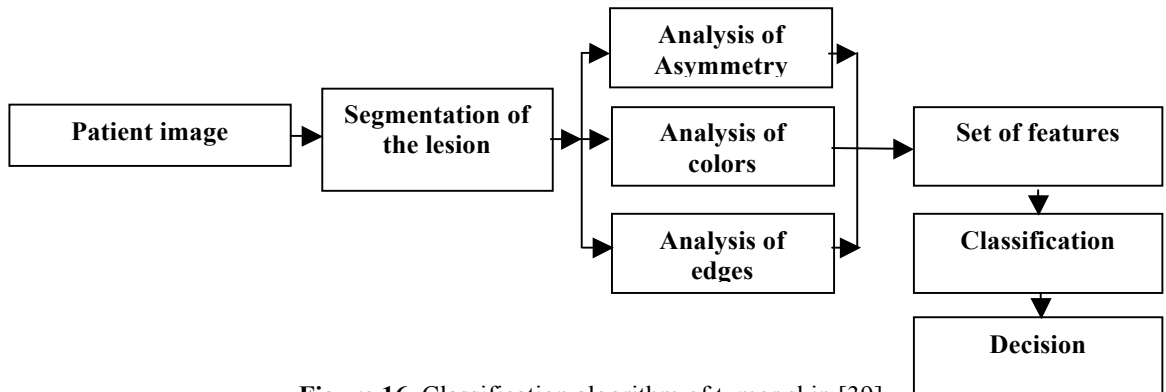

**Figure 16.** Classification algorithm of tumor skin [39].

The network architecture input' is defined by ten entry units. They represent different attributes describing the tumors (Asymmetry index, Asymmetry, Compactness, Radial variance, Irregularity index, Correlation, Colors homogeneity, Energy, Contrast and Diameter). The learning algorithm of multilayer networks, known as the back-propagation algorithm, requires that the activation functions of neurons are continuous and derivable [40-41]. The classifier differentiates between benign lesions and malignant tumors. In this case, the back-propagation algorithm minimizes the squared error $\epsilon_r$ between the desired output and the input. The classification process can study and acquire experience on melanomas and benign lesions. There are a number of arbitrary parameters whose values must be defined for the network to get good performances, in particular, the number of hidden layers and the number of iterations. In our case, the errors $\epsilon_r$ are less than 0.1 and the number of iterations, which assure the convergence of the network, is equal to 100 iterations. To assess the performance of the network, we chose to use training and testing. To evaluate the classifier performance of the designed perceptron, we used the holdout method with is the simplest kind of cross validation models. This method indicates how well the classifier will do when it is asked to make new predictions for data it has not already seen. It consists in separating the images data set into two disjoint and independent sets, called the training set and the testing set. The classification function approximator fits a function using the training set only. Then, the function approximator is asked to predict the output values for the data in the testing set. The errors it makes are accumulated as before to give the mean absolute test set error, which is used to evaluate the classifier performance. On the one hand, the size of the training set should be large enough to ensure a good classification rate. On the other hand, the size of testing set should be also large enough to increase the confidence in the results (Table 2).

The obtained results are summarized in table 2 for different numbers of hidden units (n) so that the effect of architecture on the performance could be assessed. This table records correct detection rates, using the perceptron classifier on training set. Accuracy of classification on the testing set is evaluated in terms of sensitivity Sn (percentage of malignant lesions correctly classified) and specificity Sp (percentage of benign lesions correctly classified). For each (n), the network weights are initialized randomly over [0, 1] in every execution. The final results, given in table 2, are calculated as the average over a set of 100 executions. In conclusion, given the disposed database, the perceptron with two hidden layers leads to better results with correct classification rate (TCR) of 87.32%, sensitivity (Sn) of 90.34% and specificity (Sp) of 33.29%.

**Table 2** Classification results.

| N | Sn(%) | Sp(%) | TCR(%) |
|---|-------|-------|--------|
| 1 | 82.22 | 20    | 81.11  |
| 2 | 84    | 22.22 | 78     |
| 1 | 83.34 | 31.33 | 81     |
| 2 | 90,34 | 33.29 | 87.32  |

## 6. Conclusion

In this paper, we presented different algorithms (segmentation and characterization) used for classification of pigmented skin lesion from a macroscopic image. Given the acquired color image, we started with a pre-processing step based on a marching scheme technique for its ability to remove the noise. In the second step, we have introduced a new unsupervised approach segmentation algorithm for segmenting skin lesions in dermoscopy images. Experimental results on a large dataset of 320 dermoscopy images have shown that the proposed segmentation technique is capable of providing more accurate segmentation results than methods like GrowCut algorithm [17] and Mean shift approach [18]. Then, a new system for characterizing digital images of skin lesions has been presented..A sequence of transformation has been applied to extract different attributes (ABCD). Transformations series have been performed to calculate the asymmetry parameter for digital images of skin lesions. We proposed a method that integrates two algorithms (Lee et al. [10] and Clawson et al. [11]) to measure the irregularity index parameter. Furthermore, measures of the color information contained in the lesion and diameter are proposed. Finally, all these features are used to construct a classifier allowing the diagnosis to be evaluated. The results of this study are significant and quite promising for the future. Future work will be directed towards the detection of regions inside the lesion with significant coloring.

## 7. References

[1] A. Kopf, D. Rigel, R. Friedman, The rising incidence and mortality rates of malignant melanoma, J. Dermatol. Surg. Oncol. 8 (1982) 760_/761.
[2] A. Kopf, T. Saloopek, J. Slade, A. Marghood, R. Bart, Techniques of cutaneous examination for the detection of skin cancer, Cancer Supplement 75 (2) (1994) 684_/690.
[3] H. Koh, R. Lew, M. Prout, Screening for melanoma/skin cancer: theoretical and practical considerations, J. Am. Acad. Dermatol. 20 (1989) 159_/172.
[4] W.F. Dial, ABCD rule aids in preoperative diagnosis of malignant melanoma, Cosmetic Dermatol. 8 (3) (1995) 32_/34.
[5] D.S. Rigel, R.J. Friedman, The rationale of the ABCDs of early melanoma. 29 (6) (1993) 1060_/1061.
[6] J.S. Lederman, T.B. Fitzpatrick, A.J. Sober, Skin markings in the diagnosis and prognosis of cutaneous melanoma, Arch. Dermatol.120 (1984) 1449_/1452.
[7] M.M. Wick, A.J. Sober, T.B. Fitzpatrick, M.C. Mihm, A.W. Kopf, W.H. Clark, M.S. Blois, Clinical characteristics of early cutaneous melanoma, Cancer 45 (1980) 2684_/2686.
[8] R.J. Friedman, D.S. Rigel, A.W. Kopf, Early detection of malignant melanoma: the role of physician examination and self-examination of the skin, CA Cancer J. Clin. 35 (3) (1985) 130_/151.
[9] C. Grin, A. Kopf, B. Welkovich, R. Bart, M. Levenstein, Accuracy in the clinical diagnosis of melanoma, Arch. Dermatol.126 (1990) 763_/766.
[10] Lee, T. and S. Atkins, 2000. "A new approach to measure border irregularity for melanocytic lesions", Spie 2000 Medical Imaging, San Diego.
[11] K.M. Clawson, P.J. Morrow, B.W. Scotney, D.J. McKenna, "Determination Of Optimal Axes For Skin Lesion Asymmetry Quantification", 1-4244-1437-7/07/2007 IEEE, 2007.
[12] Mariam A.Sheha, Mai S.Mabrouk, AmrSharawy, March 2012. ―Automatic Detection of Melanoma Skin Cancer usingTexture Analysis‖ International Journal of Computer Applications.
[13] Mai S. Mabrouk,Mariam A. Sheha, Amr A. Sharawy, 2013, "Computer Aided Diagnosis of Melanoma Skin Cancer using Clinical Photographic images" INTERNATIONAL JOURNAL OF COMPUTERS & TECHNOLOGY.
[14] Pablo G. Cavalcanti and Jacob Scharcanski, 2013 "Macroscopic Pigmented Skin Lesion Segmentation and Its Influence on Lesion Classification and Diagnosis" Springer Science+Business Media Dordrecht

[15] Nadia Smaoui, SouhirBessassi, 2013 " A developed system for melanoma diagnosis", International Journal of Computer Vision and Signal Processing, 3(1), 10-17
[16] Dermatologic Image Database, University of Auckland, New Zealand. Available from: http://dermnetnz.org/doctors/dermoscopy-course/ (accessed on 15.06.2009).
[17] Vezhnevets.V and V Konouchine. V, 2011 ""GrowCut" - Interactive Multi-Label N-D Image Segmentation By Cellular Automata", Graphics and Media Laboratory, Moscow, Russia.
[18] P. G. Cavalcanti and J. Scharcanski, 2013,"Macroscopic Pigmented Skin Lesion Segmentation and Its Influence on Lesion Classification and Diagnosis", Springer Science, Business Media Dordrecht 2013
[19] Cavalcanti PG, Scharcanski J (2011) Automated prescreening of pigmented skin lesions using standard cameras. Comput Med Imaging Graph 35(6):481–491
[20] Cavalcanti PG, Scharcanski J, Persia LED, Milone DH (2011) An ica-based method for the segmentation of pigmented skin lesions in macroscopic images. In: Proceedings of the 33rd annual international conference of the IEEE engineering in medicine and biology society (EMBC)
[21] Q. Abbasa, M.E. Celebic, I. FondónGarcíad, "Hair removal methods: A comparative study for dermoscopy images", Biomedical Signal Processing and Control 6 (2011) 395– 404
[22] Q. Abbas, I. Fondon, M. Rashid, Unsupervised skin lesions border detection via two-dimensional image analysis, Comput. Meth. Prog. Bio. (2010).
[23] F. Bornemann, T. März, Fast image inpainting based on coherence transport, J. Math. Imaging Vis. 28 (2007) 259–278.
[24] E. Čuk, M. Gams, M. Možek "Supervised Visual System for Recognition of Erythema Migrans, an Early Skin Manifestation of Lyme Borreliosis", Journal of Mechanical Engineering, 2013
[25] M. Sonka, V. Hlavac and R. Boyle, "Image processing, analysis, and machine vision", Third edition, Thomson, USA, 2008.
[26] C.Xu. J.I. Prince, Snakes shapes and gradient vector flow. IEEE transactions on image processing 7(1998) 359-369.
[27] V. Caselles. R. Kimmel. G. Sapiro. Geodesic active contours. International journal of Computer Vision 22 (1997) 61-79.
[28] Bing Nan Li, Chee Kong Chui, Stephen Chang, S.H. Ong. Integrating spatial fuzzy clustering with level set methods for automated medical image segmentation. Computers in Biology and Medicine 41 (2011) 1-10.
[29] Messadi, M., Bessadi, A., and taleb-ahmed, A.,2010, "Segmentation of dermatoscopic images used for computer-aided diagnosis of melanoma", Journal of Mechanics in Medicine and Biology, Vol. 10, No. 2 (2010) 1–11.
[30] Garnavi R., Aldeen M., Celebi E., "Border detection in dermoscopy images using hybrid thresholding on optimized color channels » Computerized Medical Imaging and Graphics 35 (2011) 105–115
[31] Ganster, H., Pinz, A, Rohrer, E., Wilding, E., Binder, M. and Kittler, H., 2001, Automated melanoma recognition. IEEE Transactions on Medical Imaging, 20, 233–239.
[32] Stoecker, W., Weiling, V., Li, W. and Moss, R., 1992, Automatic detection of asymmetry in skin tumors. Computerized MedicalImaging and Graphics, 16, 191–197.
[33] Chaudhuri, B., Sarkar, N., 1995, "Texture segmentation using fractal dimension", IEEE Transactions on Pattern Analysis and Machine Intelligence 17(1), p. 72-77.
[34] Ng, V., and Lee, T., 1996, "Measuring border irregularities of skin lesions using fractal dimensions", In SPIE Photonics, Electronic Imaging and Multimedia Systems, China, p. 64-72.
[35] E. Zagrouba and all "A primary approach for the automated recognition of malignant melanoma", department des sciences de l'informatique, laboratory LIP2, TST, TUNISIA, image anl stereol23-121-135, 2004.
[36] Xiaojing Yuan, Zhenyu Yang, George Zouridakis, and NizarMullani , 2006, "SVM-based Texture Classification and Application to Early Melanoma Detection", Proceedings of the 28th IEEE EMBS Annual International Conference New York City, USA, Aug 30-Sept 3, 2006, 1-4244-0033-3/06
[37] Haralick, R. M., Shanmugan, K., and Dinstein, 1973. "Textural features for image classification", IEEE Trans. Syst. Man Cybernetics SMC-3, 610 ,1973.
[38] Cucchiara, R., Grana, C. and Piccardi, M., 2001, Iterative fuzzy clustering for detecting regions of interest in skin lesions, University of Modena – 41100 Modena, Italy.
[39] Schindewolf, T., Stolz, W., et al., 1993, Classification of melanocytic lesions with color and texture analysis using digital image processing. Analytical & Quantitative Cytology & Histology, 15, 1–11.
[40] Rosenblatt, F., 1962, Principles of Neurodynamics: Perceptron and theory of brain mechanisms, Washington, DC: Spartan Books.
[41] Bostock, R., Claridge, T., Harget, E. and Hall, P., 1993, Towards a neural network based system for skin cancer diagnosis. 3rd International Conference on Artificial Neural Networks, 372, 215–219.